%% file: main.tex
\documentclass{article}
\usepackage{spconf,amsmath,graphicx}
\usepackage{graphicx}
\usepackage{amsmath}
\usepackage{amssymb}
\usepackage{booktabs}

\usepackage{algorithm}
\usepackage{algorithmic}

\usepackage{multirow}
\usepackage{multicol}
\usepackage{pifont}

\usepackage{verbatim}
\usepackage{kotex}
\usepackage{colortbl}
\usepackage[table]{xcolor}
\usepackage{bm}
\usepackage{color}
\usepackage[export]{adjustbox}
\usepackage{bibspacing}
\newcommand{\myparagraph}[1]{\vspace{5pt}\noindent{\bf #1}}

\usepackage[hidelinks]{hyperref}
\title{ViTA-PAR: Visual and Textual Attribute Alignment \\with Attribute Prompting for Pedestrian Attribute Recognition}
\name{Minjeong Park \qquad Hongbeen Park \qquad Jinkyu Kim*
\thanks{* Corresponding author: J. Kim (jinkyukim@korea.ac.kr)}
 }
\address{Department of Computer Science and Engineering, Korea University, Seoul 02841, Korea}

\begin{document}
%\ninept
%
\maketitle
\begin{abstract}
\input{0_Abstract}
\end{abstract}
\section{Introduction}
\label{sec:intro}
\input{1_Introduction}

\section{Related Work}
\vspace{-1em}
\label{sec:relatedworks}
\input{2_Related}
\vspace{-0.5em}

\section{Method}
\label{sec:method}
\input{3_Approach}

\section{Experiments}
\label{sec:experiments}
\input{4_Experiment}

\vspace{-1em}
\section{Conclusion}
\vspace{-5pt}
\label{sec:conclusion}
\input{5_Conclusion}

{
\small
\bibliographystyle{IEEEbib}
\bibliography{main}
}
\end{document}

%% file: 0_Abstract.tex
The Pedestrian Attribute Recognition (PAR) task aims to identify various detailed attributes of an individual, such as clothing, accessories, and gender. To enhance PAR performance, a model must capture features ranging from coarse-grained global attributes (e.g., for identifying gender) to fine-grained local details (e.g., for recognizing accessories) that may appear in diverse regions. Recent research suggests that body part representation can enhance the model's robustness and accuracy, but these methods are often restricted to attribute classes within fixed horizontal regions, leading to degraded performance when attributes appear in varying or unexpected body locations. In this paper, we propose \textbf{Vi}sual and \textbf{T}extual \textbf{A}ttribute Alignment with Attribute Prompting for \textbf{P}edestrian \textbf{A}ttribute \textbf{R}ecognition, dubbed as \textbf{ViTA-PAR}, to enhance attribute recognition through specialized multimodal prompting and vision-language alignment. We introduce visual attribute prompts that capture global-to-local semantics, enabling diverse attribute representations. To enrich textual embeddings, we design a learnable prompt template, termed person and attribute context prompting, to learn person and attributes context. Finally, we align visual and textual attribute features for effective fusion. ViTA-PAR is validated on four PAR benchmarks, achieving competitive performance with efficient inference. 
We release our code and model at \href{https://github.com/mlnjeongpark/ViTA-PAR}{\texttt{https://github.com/mlnjeongpark/ViTA-PAR}}.

%% file: 1_Introduction.tex
\begin{figure*}[t]
    \centering
    \includegraphics[width=0.85\linewidth]{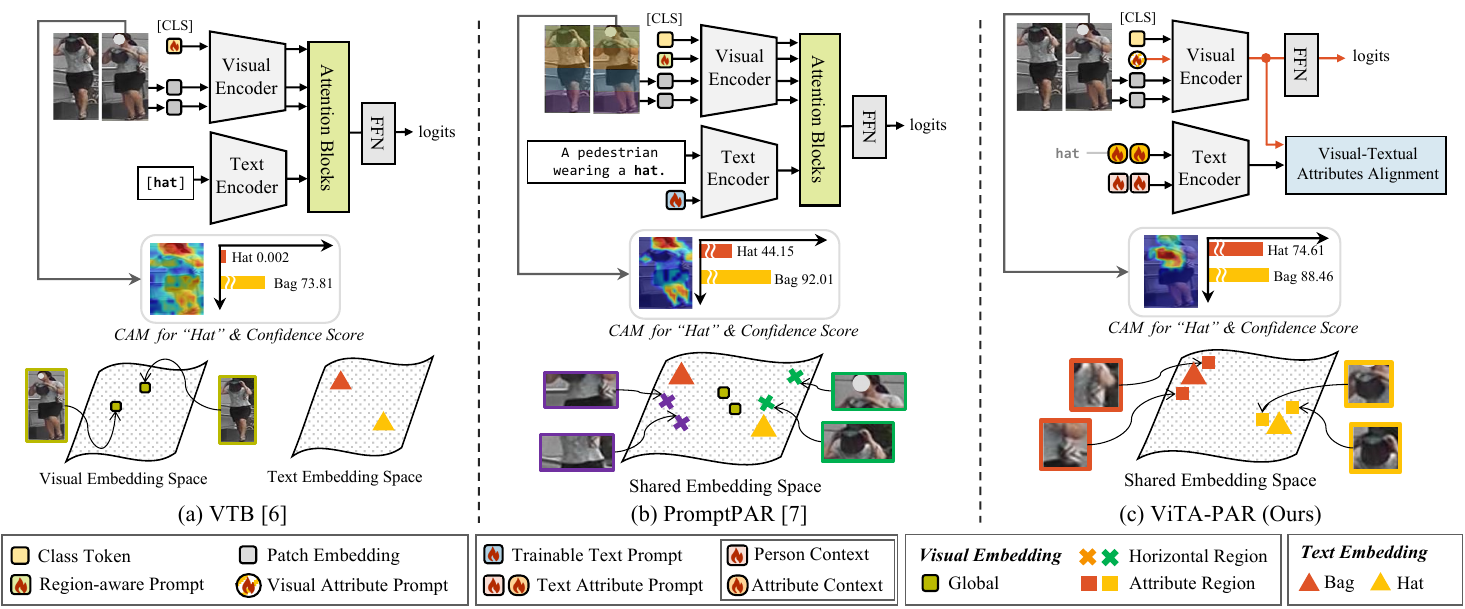}
    \vspace{-0.8em}
    \caption{Comparison of ViTA-PAR with other visual and language-based PAR (VL-PAR) approaches: VTB~\cite{vtb} and PromptPAR~\cite{wang2023promptPAR}. (a) VTB uses ViT and BERT as feature extractors, resulting in separate representation spaces that hinder direct modality fusion. Moreover, the class token, trained with long-range attention, struggles to capture fine-grained attributes, as shown in the CAM visualization of ``Hat". (b) PromptPAR leverages CLIP’s shared representation space and optimizes region-aware prompts by horizontally dividing image patches. However, due to the diverse attribute locations, the region prompt misaligns with textual embeddings, leading to suboptimal ``Hat" predictions.  (c) To address these issues, we propose visual and textual attribute prompts with visual-textual alignment. Our approach effectively captures ``hat" despite its dynamic location, as evidenced by CAM results and a higher confidence score than (a) and (b).
    }
    \vspace{-1em}
    \label{fig:teaser_figure}
\end{figure*}

Pedestrian attribute recognition (PAR) models are trained to recognize diverse visual attributes of pedestrians, such as age, gender, clothing, and personal belongings. Such PAR models play a critical role in various computer vision applications, including intelligent video surveillance~\cite{chen2022pedestrian}, person retrieval~\cite{yang2023towards, jeong2021asmr},  
and person re-identification~\cite{zhai2024multi, huang2024attribute}. 
% zhu2023attribute, lin2019improving,
Conventional PAR models are often image-based methods and formulated as a multi-label classification task, but recent success (called VTB~\cite{vtb}) suggested that visual and language based PAR (VL-PAR) can improve the overall PAR performance by integrating a visual class token (which contains the encoded features of the entire image) and textual semantics from attribute annotation (e.g., ``bag'') by a BERT-based pre-trained textual encoder (see Figure~\ref{fig:teaser_figure} (a)). Though promising, its dependency on a (global) class token often limits the model's capability to identify local and fine-grained visual attributes as shown in Figure~\ref{fig:teaser_figure}. Furthermore, this approach relies on pre-trained ViT and a BERT-based model separately, which makes it challenging to effectively exploit the internal relationships between visual and textual features due to the misalignment between their respective embedding spaces.  

More recently, CLIP~\cite{CLIP} has been employed as a feature extractor in PAR models to leverage its advantage of shared visual and textual representation space. PromptPAR~\cite{wang2023promptPAR} proposes region-aware prompts, trained by horizontally dividing image patches into four parts and by aligning these prompts with the attribute classes. Additionally, it introduces language descriptions of attributes and trainable text prompts for text embedding to fine-tune the CLIP model for PAR (see Figure~\ref{fig:teaser_figure} (b)). It alleviates the limitation mentioned above and achieves better results compared to previous works, however, its performance is hindered for the following two reasons. (1) Although certain attribute classes correspond to specific horizontal regions of the human body, their spatial alignment often deviates from expected locations. For instance, as shown in the input images in Figure~\ref{fig:teaser_figure},  a person may wear a hat and later remove it, holding it in front, causing the hat to appear in different regions, such as the head or chest. As a result, human regions do not always align reliably with attribute classes, leading to misalignment in some cases, as illustrated by the CAM visualizations and confidence scores. (2) The study of text prompt remains relatively underexplored though text prompting methods have significantly impacted the performance in various vision tasks~\cite{zhou2022coop,zhou2022cocoop,zhou2024anomalyclip}.

To capture coarse-to-fine attributes that are dynamically located in the image and make full use of pre-trained knowledge of CLIP in PAR, this paper presents a \textbf{Vi}sual and \textbf{T}extual Attribute \textbf{A}lignment for \textbf{P}edestrian \textbf{A}ttribute \textbf{R}ecognition framework, dubbed as \textbf{ViTA-PAR} as shown in Figure~\ref{fig:teaser_figure} (c). To learn global to local attribute cues from the images, we introduce a visual attribute prompt to encapsulate specialized visual semantic representations of each attribute. Subsequently, we propose a simple yet effective learnable text prompt template called person and attribute context prompting to provide a more comprehensive context of person and attributes, and it is adopted for text attribute prompts to learn detailed person and attribute characteristics, effectively covering a broad spectrum of attribute semantics. For visual-textual attribute alignment, cosine similarity between the visual and textual features is calculated, which eliminates the need for additional attention blocks typically required in conventional VL-PAR approaches. During inference, we use only visual attribute features, achieving a balance between accuracy and efficiency. Based on this design, our model achieves competitive performance while offering improved inference speed compared to other VL-PAR methods as shown in Table~\ref{tab_inference-time}. 

We summarize our contributions as follows:
\begin{itemize}
    \setlength\itemsep{-0.5em}
    \item We propose a novel approach to pedestrian attribute recognition (PAR), termed ViTA-PAR, which introduces visual and textual attribute prompts and aligns their features through visual-textual attribute alignment to effectively encapsulate multimodal attribute semantics. 

    \item The person and attribute context prompting is introduced as a novel learnable text prompt template designed to capture detailed characteristics of both the person and the attributes. Its learnable prompt design enables the model to capture diverse attribute patterns effectively.

    \item Extensive experiments demonstrate that ViTA-PAR outperforms existing approaches while offering a more efficient inference time, making it well-suited for real-world applications. 

\end{itemize}
\vspace{-1em}

%% file: 2_Related.tex
\myparagraph{Pedestrian Attribute Recognition (PAR).}
Early PAR approaches formulated the PAR problem as a multi-label classification.
%, where visual features are extracted from convolutional neural networks (CNNs) followed by a classifier built upon graph convolutional networks (GCNs)~\cite{2020JLAC} and recurrent neural networks (RNNs)~\cite{wang2017attribute, zhao2018grouping}.
Recently, VTB~\cite{vtb} leveraged language modality to utilize textual semantics in PAR tasks, improving the model's representation and generalization power. Though promising, two limitations remain: (i) textual embeddings were not well-aligned with visual embeddings, and (ii) their dependency on a class token of ViT limits the model to focus on local fine-grained attributes, achieving sub-optimal PAR performance. PromptPAR~\cite{wang2023promptPAR} addressed (i) by utilizing CLIP-based image-text joint embedding space and (ii) by region-aware prompts, where each prompt is assigned to a fixed horizontally divided image region (e.g., a prompt is simply assigned to the top part of an image). However, we argue that a better PAR performance can be achieved by (i) proposing attribute-aware prompts, which are no longer constrained by fixed image regions, and (2) utilizing text prompting methods to use textual semantics better. 

\myparagraph{Prompt Learning.}
Visual-language models (VLMs), such as CLIP~\cite{CLIP}, have widely been applied to various computer vision tasks~\cite{zhou2024anomalyclip, li2023clipreid} because of their robust capability to understand and correlate textual and visual information. Though promising, as mentioned in \cite{CLIP}, they often struggle with recognizing local fine-grained attributes. Recent studies suggest various prompt learning methods~\cite{zhou2022coop, zhou2022cocoop}. 
While they could provide more detailed descriptions than engineered prompts and represent the overall characteristics of the image, they fall short of fully describing the characteristics of each attribute class, as they simply insert the class name into the prompt. This approach performs well in capturing global foreground objects but may lack the capability to capture properties specific to local regions accurately. Thus, in this work, we construct a novel person and attribute context prompting. This prompt design, which includes person and attribute contexts, enables the model to learn the overall features of the image as well as the features unique to each attribute.

%% file: 3_Approach.tex
\vspace{-0.5em}
\begin{figure*}[!t]
    \centering
    \includegraphics[width=0.9\linewidth]{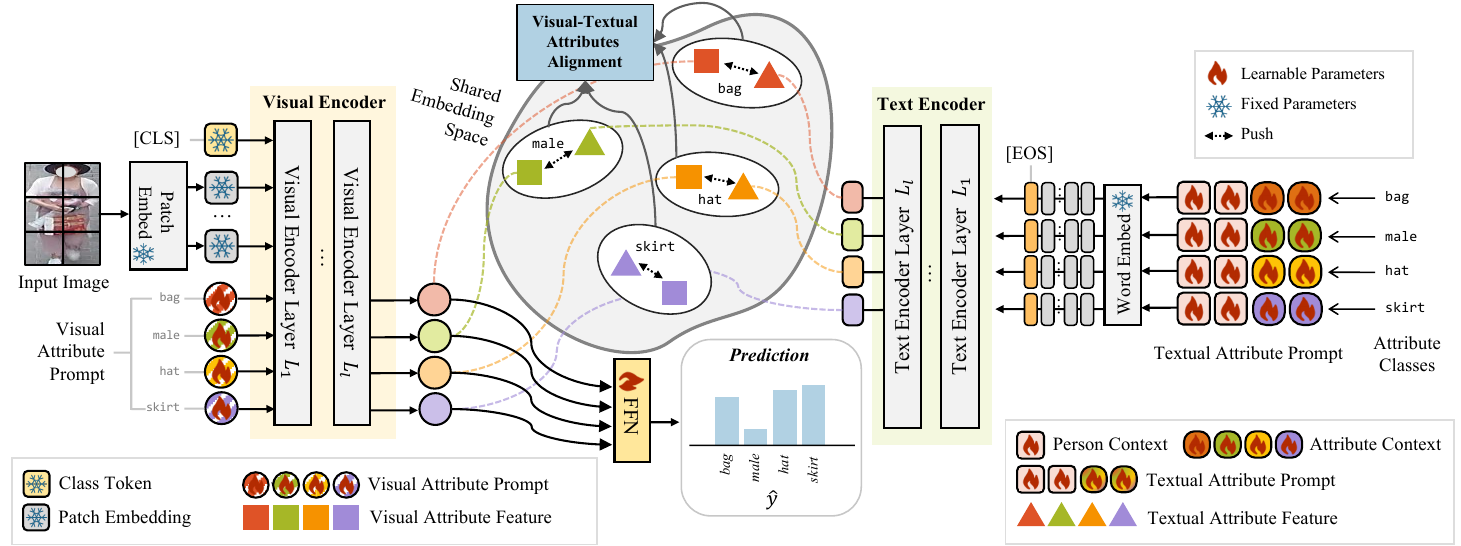}
    \caption{\textbf{ViTA-PAR Architecture.} 
    ViTA-PAR introduces visual and textual attribute prompts to capture coarse-to-fine attribute cues and align them within a shared embedding space through visual-textual attribute alignment. Note that, during testing, only the image is fed into the visual encoder, excluding textual features, which leads to reduced computational cost.
    }
    \label{fig:main_figure}
    \vspace{-1em}
\end{figure*}

The whole pipeline of the ViTA-PAR is shown in Figure~\ref{fig:main_figure}. We adopt CLIP as a feature extractor. We introduce visual attribute prompt $P^v \in \mathbb{R}^{A \times d}$ to capture coarse-to-fine visual cues and text attribute prompts $P^t \in \mathbb{R}^{A \times d}$ generated by the person and attribute context prompting to cover diverse description of person and attribute.
% and the prompts are fed into the text encoder $\mathcal{T}$. 
We design a visual-text attribute alignment mechanism that aligns each visual attribute embedding with its corresponding textual attribute embedding, enabling the visual attribute features to encapsulate diverse attribute semantics derived from the text. During inference, we utilize only image features, achieving improved computational efficiency while maintaining superior performance. More details are described in the following sections.

\vspace{-1em}
\subsection{Visual Attribute Prompt}
The conventional class token ([CLS]) from ViT effectively identifies long-range relations between multiple image patches, capturing global contextual information. This makes it particularly suitable for recognizing attributes like age and gender, which need to understand global semantic representations across a broader region of a given image. However, attributes such as clothing and accessories are often localized, making it challenging to capture their fine-grained local semantic representation from the globally focusing ViT class token. To address this, we introduce a visual attribute prompt to capture global and local visual representations of attribute classes. To leverage CLIP's pre-trained knowledge of visual embedding, we initialize these parameters with the pre-trained weight of a class token. The visual attribute prompt and the input image are taken as the input of the visual encoder $\mathcal{V}$. Specifically, given an input image $I \in \mathbb{R}^{H \times W \times 3}$, $I$ is divided into fixed-size patches and projected into patch embedding. A class token $c_0 \in \mathbb{R}^{d_v}$, embedded patches $E_0 \in \mathbb{R}^{P_v \times d_v}$ and visual attribute prompt $P^v_0 \in \mathbb{R}^{A \times d_v}$ are concatenated together and sequentially processed through $K$ visual encoder blocks, where $A$ is the number of attribute classes and $d_v$ is channel dimension.
\vspace{-0.5em}
\begin{equation}
[c_i, E_i, P_i^v] = \mathcal{V}_i([c_{i-1}, E_{i-1}, P_{i-1}^v]) \quad i=1,2,...,K
\end{equation}
where $[\cdot ,\cdot]$ is concatenate operation. The visual attribute prompt $P^v_K$ from the last visual encoder layer is processed in two ways. First, it is projected to a shared embedding space and generates visual attribute features using the VisProj layer, following the approach used in CLIP as expressed as Eq.~\ref{eq:VisProj}. 
\vspace{-0.2em}
\begin{equation}
\label{eq:VisProj}
    f_v = \textrm{VisProj}(P_K^v) \qquad f_v\in \mathbb{R}^{A \times d_{vt}}
\end{equation}
Second, it is fed into a feed-forward layer (FFN) for attribute prediction, which provides a balance of computational efficiency and strong performance. We employ a weighted cross-entropy loss function \cite{deepmar_rap} to adjust the imbalanced data distribution: 
\vspace{-0.2em}
\begin{equation}
\label{eq:Pred}
    \hat{y} = \textrm{sigmoid}(\textrm{FFN}(P_K^v))
    \qquad 
    \hat{y} \in \mathbb{R}^{A}
\end{equation}
\vspace{-0.54cm}
\begin{equation}
\mathcal{L}_{pred}=\sum_{j=1}^{A} w^j(y^j\log(\hat{y}^j)+(1-y^j)\log(1-\hat{y}^j))
\end{equation}
where $w^j$ is the imbalance weight of the $j$-th attribute.
\vspace{-5pt}
\subsection{Contextual Templates for Text Prompt}
\vspace{-5pt}
\myparagraph{Person and Attribute Context Prompting.}
Mostly used text prompt engineering and learning design, such as `A photo of a [class]' or `$[X]_1[X]_2[X]_3...[X]_M$ [class]', where $[X]_m (m \in 1, ..., M)$ is a learnable parameter and $M$ is the number of learnable context tokens, primarily focus on describing foreground classes in the image. However, certain attribute classes (e.g., accessories, shoes) may not always be the primary focus of the image. A straightforward approach to constructing sentences for coarse-to-fine-grained attributes is to incorporate detailed descriptions directly into the prompts. For instance, for the attribute `backpack', the prompt could be structured as `A photo of \textit{a person with} the [backpack] \textit{on their back}' or `$[X]_1[X]_2[X]_3...[X]_M$ [backpack] \textit{on their back}'. However,  listing all possible descriptions is challenging due to the diverse patterns of attributes, highlighting the necessity for a more flexible and adaptive approach to prompt design. To address this, we have developed a new learnable prompt template, termed person and attribute context prompting. The template consists of learnable tokens representing both person and attribute contexts. Specifically, the person context, i.e., $[X]_1[X]_2...[X]_G$, shared across all attribute classes, is designed to learn a general representation of a person in an image, while the attribute context, i.e., $[Y]_1[Y]_2...[Y]_S$ is unique to each class, replacing the attribute class name in the conventional text prompt and capturing the distinct and comprehensive patterns of the attributes. The prompt used to describe a specific attribute class is defined as follows:
\vspace{-0.5em}
\begin{equation}
t = [X]_1[X]_2...[X]_G [Y]_1[Y]_2...[Y]_S \quad t \in \mathbb{R}^{(G+S) \times d_t}
\end{equation}
where $[X]_p$ $(p \in 1,...,G)$ and $[Y]_q$ $(q \in 1,...,S)$ are a learnable $d_t$-dimensional vector, and $G$ and $S$ are the lengths of the person and attribute context, respectively. 

\myparagraph{Textual Attribute Prompt.}
The person and attribute context prompting is adopted to generate the textual attribute prompts as follows:
\vspace{-0.5em}
\begin{equation}
P^t = \{[X]_1...[X]_G [Y]_1^j...[Y]_S^j\}_{j=1}^A \quad P^t \in \mathbb{R}^{A\times (G+S) \times d_t}
\end{equation}
The prompts $P^t$ are tokenized, and a tokenized attributes sequence $T_0=[T_0^1, T_0^2, \dots, T_0^A] \in \mathbb{R}^{A\times L\times d_t}$ is generated. The max sequence length of each attribute sequence is limited to $L$ and bracketed by [SOS] and [EOS] tokens. The tokenized sequence is forwarded to $K$ text encoder layers as follows,
\vspace{-0.5em}
\begin{equation}
T_i = \mathcal{T}_i(T_{i-1}) \quad i=1,2,...,K
\end{equation}
Finally, the [EOS] tokens $z_K \in \mathbb{R}^{A \times d_t}$ of the last transformer layer's sequence $T_K$ are projected to the shared embedding space by the TextProj layer to obtain the textual attribute features,

\vspace{-0.8em}
\begin{equation}
f_t = \textrm{TextProj}(z_K) \qquad z_k\in \mathbb{R}^{A \times d_{vt}}
\end{equation}
\vspace{-2.5em}
\subsection{Visual-Textual Attributes Alignment}
We reason that aligning visual and textual features is crucial for efficiently leveraging CLIP's pre-trained knowledge for PAR rather than attention-based fusion, given its pre-training strategy. Furthermore, we demonstrate that attribute prediction is more effective even in the absence of textual features, as long as the alignment between textual and visual features is well-established in Table ~\ref{tab_ablation}. To the end, we propose a visual-textual attribute alignment method to calculate the cosine similarity ($sim(\cdot,\cdot)$) with a temperature parameter $\tau$ between the visual and textual attribute features. The similarity scores are normalized and forwarded to the Sigmoid function for prediction scores. The binary-cross entropy loss is applied between the prediction scores $\hat{y}_{vt}$ and ground truths $y$:
\vspace{-0.5em}
\begin{equation}
\hat{y}_{vt} = \textrm{sigmoid}(sim(f_v, f_t)/\tau) \qquad \hat{y}_{vt} \in \mathbb{R}^{A} 
\end{equation}
\vspace{-1.5em}
\begin{equation}
\mathcal{L}_{align}=\sum_{j=1}^{A} y^j\log(\hat{y}_{vt}^j)+(1-y^j)\log(1-\hat{y}_{vt}^j)
\end{equation}

The training losses are summarized as follows:
\vspace{-0.5em}
\begin{equation}
\mathcal{L}=\alpha\mathcal{L}_{pred}+\beta\mathcal{L}_{align}
\end{equation}
\vspace{-2.5em}

%% file: 4_Experiment.tex
\vspace{-1em}
\myparagraph{Datasets.}
We conducted experiments on four publicly available datasets—PA-100K~\cite{PA100k}, PETA~\cite{PETA}, RAPv1~\cite{rap1}, and RAPv2~\cite{rap2}—covering indoor and outdoor scenarios to evaluate the performance of ViTA-PAR. We follow the standard dataset protocol \cite{jia2021rethinking} to ensure consistency and reproducibility in our experiments.

\input{tables/4_1_SOTA}
\input{tables/4_2_inferencetime}

\myparagraph{Implementation and Evaluation Details.}
We adopt CLIP with ViT-B/16 and ViT-L/14 as the visual encoder and keep the same CLIP Transformer as the text encoder. We freeze the text encoder and set the image size to $256 \times 192$. The model is trained for 100 epochs with a batch size of 32 on an A6000 GPU. The parameters are optimized using the AdamW optimizer with an initial learning rate of 2e-3, decayed the cosine annealing schedule. We chose $\alpha =1, \beta=0$ for the first 10 epochs, $\alpha =0, \beta=1$ for the next 10 epochs, and $\alpha =1, \beta=0.5$ for the remaining epochs. $G, S$ are set to 4, 16. For evaluation metrics, we used two widely recognized measures: mean accuracy (mA) and the F1-score.

\myparagraph{Quantitative Analysis.}
We compare proposed ViTA-PAR with the state-of-the-art methods on four PAR benchmarks as shown in Table~\ref{tab_SOTA}. ViTA-PAR demonstrates superior performance compared to both image-based and visual-language-based methods on the three datasets. On PETA, ViTA-PAR achieves the second-best performance in terms of mA, while attaining the best performance in F1-score. Even though our model does not use textual features during inference, it achieves higher mA and F1 scores compared to other methods that utilize the CLIP feature extractor. Inference times are reported in milliseconds (msec) as shown in Table~\ref{tab_inference-time}. We experimented on four samples using a single A6000 GPU and compared ViTA-PAR with the methods utilizing CLIP as the feature extractor. ViTA-PAR on ViT-L/14 achieves a 2.25× speedup over PromptPAR and a 1.60× speedup over VTB, corresponding to a reduction in inference time of 55.65\% and 37.41\%, respectively. On ViT-B/16, ViT-L/16 shows 5.22×, and 3.70× speedup over PromptPAR and VTB while achieving competitive performance.

\begin{figure}[t]
    \centering
    \includegraphics[width=0.9\linewidth]{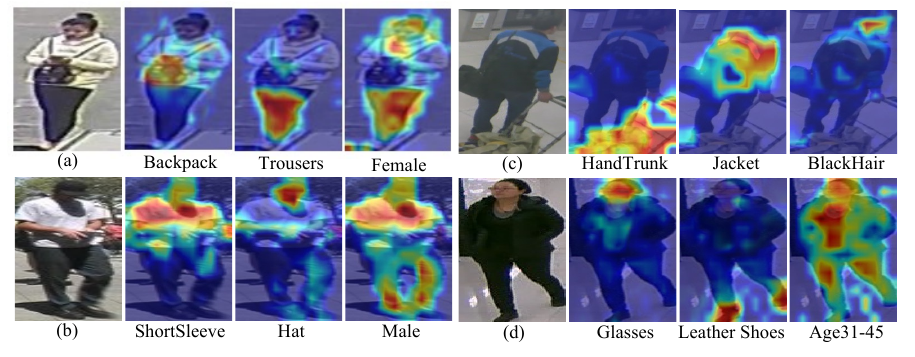}
    \vspace{-1.4em}
    \caption{Class activation maps for randomly selected images from (a) PA-100K, (b) PETA, (c) RAPv1, and (d) RAPv2 datasets.} 
    \label{fig:cam}
    \vspace{-1.5em}
\end{figure}

\myparagraph{Qualitative Analysis.}
We present the class activation maps across the datasets in Figure~\ref{fig:cam}. The heatmaps visualize the regions of interest for each attribute (e.g., backpack, trousers, female, etc.) across the datasets, highlighting the most relevant areas to attribute detection (see our method correctly identifies various pedestrian attributes across different datasets). Heat maps are color-coded, with red indicating a high response and blue indicating a low response. The backpack in the image in Figure~\ref{fig:cam} (a) is located in front of the person, which deviates from the typical placement. Despite this atypical configuration, ViTA-PAR successfully identified and highlighted the region, demonstrating its robust ability to adapt to and accurately capture diverse attribute patterns. ViTA-PAR consistently exhibits a high level of confidence in the regions associated with relevant attributes, regardless of whether these attributes are situated in global or local semantic areas.
\input{tables/4_3_ablation}

\myparagraph{Ablation Study.}
In Table~\ref{tab_ablation}, we present the effectiveness of learnable attribute prompts. We keep the CLIP visual encoder fixed for the fist row method and tune only the FFN layer. We also compare the performance difference between $\hat{y}_{vt}$ and $\hat{y}$, and the text prompting methods in the third to the sixth row method.  `Person Context' denotes $P^t$ containing only person context, similar to previous approaches~\cite{zhou2022coop,wang2023promptPAR}. The model achieves strong performance using only visual features, highlighting the effectiveness of our alignment strategy.

%% file: tables/4_1_SOTA.tex
{
\aboverulesep = 0.3ex
\belowrulesep = 0.3ex
\begin{table*}[!t]
  \centering
  \scriptsize
  \resizebox{0.9\linewidth}{!}{
  \begin{tabular}{l|c|c|cc|cc|cc|cc}
    \toprule
\multicolumn{1}{c|}{\multirow{2}{*}{Methods}} & 
\multicolumn{1}{c|}{\multirow{2}{*}{Backbone}} & 
\multicolumn{1}{c|}{Text} & 
\multicolumn{2}{c|}{PA-100K~\cite{PA100k}} & 
\multicolumn{2}{c|}{PETA~\cite{PETA}} &
\multicolumn{2}{c|}{RAPv1~\cite{rap1}} & 
\multicolumn{2}{c}{RAPv2~\cite{rap2}} \\ 
\cmidrule{4-11}
\multicolumn{1}{c|}{} &
  \multicolumn{1}{c|}{} &
  \multicolumn{1}{c|}{Infer.} &
  \multicolumn{1}{c}{mA$\uparrow$} &
  \multicolumn{1}{c|}{F1$\uparrow$} &
  \multicolumn{1}{c}{mA$\uparrow$} &
  \multicolumn{1}{c|}{F1$\uparrow$} &
    \multicolumn{1}{c}{mA$\uparrow$} &
  \multicolumn{1}{c|}{F1$\uparrow$} &
  \multicolumn{1}{c}{mA$\uparrow$} &
  \multicolumn{1}{c}{F1$\uparrow$} \\
    \midrule
    JLAC \cite{2020JLAC} & ResNet50 & \ding{55} & 82.31 & 87.61 & 86.96  & 87.50
    & 83.69   &    80.82   &   79.23    &    77.40    \\
    DAFL \cite{dafl} & ResNet50 & \ding{55} & 83.54 & 88.09 &  87.07 & 86.40
    & 83.72   &   80.29    &   81.04    &   79.13    \\
    FEMDAR \cite{cao2023novel} & ResNet50 & \ding{55} & 81.02 & 87.32 & 84.73 & 85.90
    & 79.71    &     78.76   &    -   &    -   \\
    DRFormer \cite{2022drformer} & ViT-B/16 & \ding{55} & 82.47 & 88.04 &  \textbf{89.96} &      88.30 
    & 81.81 & 81.42 & - & -  \\
    DFDT \cite{zheng2023diverse} & SwinT-B & \ding{55} & 83.63 & 88.74 & 87.44 &81.17
    & 82.34  &   {82.15}  &  79.96 &  {80.97} \\
    PARFormer \cite{fan2023parformer} & SwinT-L & \ding{55} & 84.46 & 88.52 & 89.32 & {89.06}
    &  84.13     &   81.35    &    -   &    -   \\
    VTB \cite{vtb} & ViT-B/16 & \ding{52} & 83.72 & 88.21 & 85.31 & 86.71 
    & 82.67     &    80.84   &    {81.34}   &   79.35    \\
    \midrule
    VTB* \cite{vtb} & ViT-L/14 & \ding{52} & 85.30  & 88.86 & 86.34 & 86.97 
    & {85.30} & 81.10 & 81.36 & 79.52\\
    PromptPAR* \cite{wang2023promptPAR} & ViT-L/14 & \ding{52} & \underline{87.47} & \underline{90.15} & 88.76 & {89.18}
    & \underline{85.45} & 82.38 & \underline{83.14} & 81.00\\\midrule
    \cellcolor{gray!20}\textbf{ViTA-PAR (Ours)} & 
    \cellcolor{gray!20} ViT-B/16 & \cellcolor{gray!20}\ding{55} & 
    \cellcolor{gray!20}85.91    &
    \cellcolor{gray!20}90.01    &
    \cellcolor{gray!20}89.19    &
    \cellcolor{gray!20}\underline{89.48}   &
    \cellcolor{gray!20} {84.99}    &
    \cellcolor{gray!20} \underline{82.79}    &
    \cellcolor{gray!20} {82.38}    &
    \cellcolor{gray!20} \underline{81.85}    \\
    \cellcolor{gray!20}\textbf{ViTA-PAR (Ours)} & 
    \cellcolor{gray!20} ViT-L/14 & 
    \cellcolor{gray!20}\ding{55} & 
    \cellcolor{gray!20} \textbf{87.82}    &
    \cellcolor{gray!20} \textbf{90.93}    &
    \cellcolor{gray!20} \underline{89.68}    &
    \cellcolor{gray!20} \textbf{89.83}    &
    \cellcolor{gray!20} \textbf{85.96}    &
    \cellcolor{gray!20} \textbf{83.65}    &
    \cellcolor{gray!20} \textbf{83.27}    &
    \cellcolor{gray!20} \textbf{82.57} \\
  \bottomrule
\end{tabular}}
\caption{Performance comparisons with the state-of-the-art methods on PA-100K, PETA, RAPv1 and RAPv2. The best performance is highlighted in \textbf{bold}, and the second highest score is highlighted in \underline{underline}.``Text Infer." refers to methods that incorporate text during inference. Methods marked with ``*" indicate that CLIP is used as the feature extractor.}
  \label{tab_SOTA}
\vspace{-0.3cm} 
\end{table*}
}

%% file: tables/4_2_inferencetime.tex
{
\aboverulesep = 0.22ex
\belowrulesep = 0.2ex
\begin{table}
    \centering
    \resizebox{\linewidth}{!}{
    \begin{tabular}{l|c|c|c|c}
        \toprule
         \multicolumn{1}{c|}{\multirow{2}{*}{Methods}} & 
          \multicolumn{1}{c|}{\multirow{2}{*}{Backbone}} & 
          \multicolumn{1}{c|}{Inference Time} & 
         \multicolumn{2}{c}{{Speedup over}}
         \\
         % \addlinespace[-0.2em]
        \cmidrule{4-5} 
         &  &  (msec) $\downarrow$ &  VTB $\uparrow$  & PromptPAR $\uparrow$ \\
        \midrule
         PromptPAR~\cite{wang2023promptPAR} & ViT-L/14 & 113.1119 & 0.91$\times$&-\\
         VTB~\cite{vtb} & ViT-L/14 & 80.1845 & - & 1.14$\times$\\
         \cellcolor{gray!20}{ViTA-PAR(Ours)} & 
                 \cellcolor{gray!20}ViT-L/14 & 
                 \cellcolor{gray!20}{{50.1853}} &
                \cellcolor{gray!20}1.60$\times$ &
                \cellcolor{gray!20}2.25$\times$
                \\
         \cellcolor{gray!20}{ViTA-PAR(Ours)} & 
                 \cellcolor{gray!20}ViT-B/16 & 
                 \cellcolor{gray!20}{{21.6572}}  &
                 \cellcolor{gray!20}3.70$\times$ &
                 \cellcolor{gray!20}5.22$\times$
                 \\
        \bottomrule
    \end{tabular}
    }
    \vspace{-1em}
    \caption{Inference time on PA-100K.}
    \vspace{-1em}
    \label{tab_inference-time}
\end{table}
}

%% file: tables/4_3_ablation.tex
{
\begin{table}
    \centering
    \resizebox{\linewidth}{!}{
    \begin{tabular}{ll|cc}
        \toprule
        Methods & & mA & F1 \\
        \midrule
        CLIP Visual Encoder  && 81.06 & 81.22 \\
        \midrule
        + Visual Attribute Prompt  && 83.67 & 88.65 \\
        \midrule
          & Person Context ($\hat{y}_{vt}$)&  87.43 & 90.51\\
        \cmidrule{2-4}
        
        + Textual Attribute Prompt & Person and Attribute Context ($\hat{y}_{vt}$) & 86.30 & 90.44 \\
        \cmidrule{2-4}
        
        & Person Context ($\hat{y}$)& 86.31 & 90.12\\
        \cmidrule{2-4}
        
        \cellcolor{gray!20}{}& \cellcolor{gray!20}{Person and Attribute Context ($\hat{y}$)} &
        \cellcolor{gray!20}\textbf{87.82} &
        \cellcolor{gray!20}\textbf{90.93} \\
        \bottomrule
    \end{tabular}}
    \caption{{Ablation study to evaluate the learnable attribute prompts on PA-100K.}}
    \label{tab_ablation}
    \vspace{-1.5em}
\end{table}
}

%% file: 5_Conclusion.tex
This paper proposes a one-to-one visual and textual alignment for the pedestrian attribute recognition framework, called ViTA-PAR. 
We introduce visual and textual attribute prompts and aligns their features through visual-textual attribute alignment to effectively encapsulate multimodal attribute semantics. The person and attribute context prompting is introduced as a novel learnable text prompt template designed to capture detailed characteristics of both the person and the attributes. We validate ViTA-PAR on four widely used datasets, and the results demonstrate the effectiveness of our approach.
\\
\\
\scriptsize{\textbf{Acknowledgement.} This work was the result of project supported by KT(Korea Telecom)- Korea University R\&D Center. This work was partly supported by Institute of Information \& communications Technology Planning \& Evaluation(IITP) under IITP grant (RS-2025-02263754, 10\%), the Leading Generative AI Human Resources Development(IITP-2025-RS-2024-00397085, 30\%) grant and ICT Creative Consilience Program (IITP-2025-RS-2020-II201819, 10\%) funded by the Korea government(MSIT).}